\documentclass[letterpaper, 10 pt, conference]{ieeeconf}  

\IEEEoverridecommandlockouts                              
\usepackage{graphics} 
\usepackage{subcaption}
\usepackage{array}
\usepackage{multirow}
\usepackage{epsfig} 
\usepackage{amsmath} 
\usepackage{algorithm}
\usepackage[noend]{algpseudocode}
\usepackage{caption}
\newcolumntype{M}[1]{>{\centering\arraybackslash}m{#1}}

\title{\LARGE \bf
Learning to infer: RL-based search for DNN primitive selection on Heterogeneous Embedded Systems
}

\author{Miguel de Prado$^{1,2}$, Nuria Pazos$^{1}$ and Luca Benini$^{2}$ \\
\parbox{6 in}{\centering 
		  $^{1}$ He-Arc Ingenierie, HES-SO. $\{$miguel.deprado, nuria.pazos$\}$@he-arc.ch \\
          $^{2}$ Integrated Systems Laboratory, ETH Zurich. lbenini@iis.ee.ethz.ch \\
}
}

\begin{document}

\maketitle
\thispagestyle{empty}
\pagestyle{empty}

\begin{abstract}

Deep Learning is increasingly being adopted by industry for computer vision applications running on embedded devices. While Convolutional Neural Networks' accuracy has achieved a mature and remarkable state, inference latency and throughput are a major concern especially when targeting low-cost and low-power embedded platforms. CNNs' inference latency may become a bottleneck for Deep Learning adoption by industry, as it is a crucial specification for many real-time processes. Furthermore, deployment of CNNs across heterogeneous platforms presents major compatibility issues due to vendor-specific technology and acceleration libraries. \par
In this work, we present QS-DNN, a fully automatic search based on Reinforcement Learning which, combined with an inference engine optimizer, efficiently explores through the design space and empirically finds the optimal combinations of libraries and primitives to speed up the inference of CNNs on heterogeneous embedded devices. We show that, an optimized combination can achieve 45x speedup in inference latency on CPU compared to a dependency-free baseline and 2x on average on GPGPU compared to the best vendor library.
Further, we demonstrate that, the quality of results and time ``to-solution'' is much better than with Random Search and achieves up to 15x better results for a short-time search.
\end{abstract}

\section{INTRODUCTION}
Artificial Intelligence (AI) is rapidly growing and will soon become ubiquitous and pervade our daily life. In particular, Deep Learning (DL) has rapidly grown in the last years achieving remarkable results in computer vision \cite{facerecognition} and speech recognition \cite{speechrecognition}. Adoption of AI by major industrial partners, e.g. Google \cite{google}, Tesla \cite{tesla}, is already a reality and its wide-range applications are to bring on a new technological revolution. \par
Convolutional Neural Networks (CNN) are one of the most successful examples of Deep Learning due to their remarkable accuracy and flexibility to many applications. CNNs are capable of learning abstract features by stacking many layers in parallel and in depth, which turns them into complex architectures. Training of CNNs has drawn great attention in the last years towards building more and more competitive and accurate architectures and surpassing human capabilities, e.g. ImageNet competition \cite{imagenet}. \par
Deployment of CNNs is not a trivial problem. However, it has not been on the focus until recently ago. The inference time, latency of the forward pass of a network, has become one of the main issues for the industrial stakeholders who would like to take up AI solutions for edge applications. Inference time represents a bottleneck in IoT or embedded devices due to the restricted resources they have and the large computational requirements \cite{anoverview}. \par
Moreover, deployment of CNNs on embedded devices presents further difficulties due to the restrictions and dependencies that the wide variety of implementations may impose in terms of frameworks e.g. Caffe \cite{caffe}, Tensorflow \cite{tensorflow}, acceleration libraries, e.g. cuDNN \cite{cudnn}, ArmCL \cite{armcl} or heterogeneous embedded platform types, e.g. CPU \cite{qualcomm}, FPGA \cite{xilinx}, GPU \cite{nvidia}. Each layer of a network may be executed by many possible libraries (and primitives from the library), or even in different processor, giving out quite a different performance. Hence, the space of approaches for CNN deployment becomes too large to test and obtain an optimal implementation \cite{andrew}, which usually results in the stakeholders selecting a single good-performing library. \par
The objective of this work is to ease the deployment of CNNs for industrial applications on a wide range of embedded platforms and automatically search for the best primitive combination to speed up the performance. To fulfill this objective, we present QS-DNN, an automatic exploration framework, which relies on a design space search based on Reinforcement Learning \cite{rl_1}. The RL-based search efficiently explores through the design space and finds an optimized combination of primitives that can be used to execute inference for a target CNN on a given platform. The search is combined with an inference engine optimizer which enables the production and deployment of CNN implementations for heterogeneous platforms. Thereby, we are able to obtain an optimized implementation by directly acquiring empirical measurements on embedded AI applications and notably boosting the performance of the process. \par
We demonstrate the effectiveness of the method by applying it to several types of CNNs for image classification, face recognition and object detection on a heterogeneous platform. On average, we achieve 2x speedup in inference latency in the ImageNet benchmark, compared to the best vendor library on a GPGPU platform. The runtime of our RL-based optimized is also very reasonable: 5 minutes are sufficient to find solutions that consistently outperform those found by Random Search with the same time budget. \par

The paper is organized as follows: in Section 2, the State-of-the-Art is presented. Section 3 describes the inference of a DNN on heterogeneous devices and the inference engine optimizer. In Section 4, we address the problem of primitive selection and describe Reinforcement Learning. In Section 5, we introduce the RL-based search engine and the methodology of the experiments. Section 6 presents the results and discussion.

\section{RELATED WORK}
We find two main topics related to this work: Auto-tuning and Machine Learning for Design Space Exploration. \par
\textbf{Auto-tuning.} 
The massive computation that CNNs demand prompts for several optimization approaches for inference on embedded devices. We can categorize two main classes: \textit{i)} computational graph engines and \textit{ii)} acceleration libraries for specific layers. 
Computation graph engines reduce execution time and memory footprint by removing overhead dependencies, fusing pipelined operations and performing cross-layer optimizations \cite{tensorflow}. Moskewicz et al. \cite{boda} use meta-programming and auto-tuning to provide portable implementations across different GPU-vendor platforms. However, their auto-tuning process is inefficiently done as they use brute force to search through the design space. Truong et al. \cite{latte} implemented Latte, a domain-specific language that abstracts the architecture and computation of a neural network. Latte's compiler is able to recognize dependencies and match patterns to perform cross-layer optimization as well as optimized-primitive calls.\par
In this work, we rather focus on the second approach: acceleration libraries for specific layers. We leverage primitives from acceleration libraries to speed up the performance of standard neural network layers. We draw inspiration from Anderson et al. \cite{andrew} who use PBQP to optimize inference time by selecting suitable backends. In their work, they profile each implementation type and the transformation cost between different implementations. They make an optimization problem to select the best backend per layer which they solved with PBQP. However, they only profile convolutional layers and do not optimize any other layer type. In addition, we propose a totally different search method which is modeled as a learning problem and implements a sample-based approach. Our method drastically reduces the space exploration effort, while still obtaining an optimized solution. \\
\par
\textbf{Machine Learning (ML) for Design Space Exploration (DSE).} 
General ML techniques have been applied as an automatic-search tool for large space exploration problems such as performance of processors \cite{processor} or high-level synthesis \cite{synthesis}.
Lately, there has been an increasing trend of using Reinforcement Learning (RL) and Evolutionary Algorithms (EA) to build CNN architectures. EA works like \cite{ea_1}, \cite{ea_2}, \cite{ea_3} use Genetic Algorithms over a population of CNNs. By using mutation operators, the architectures of the population evolve towards new topologies. Baker et al. \cite{rl} used RL to sequentially choose CNN layers. They used Q-learning employing an $\epsilon$-greedy strategy, which trades off exploration and exploitation. All these works share the assumption of fixed-sized number of parameters to select from (RL) or from which they can mutate (EA). However, they only take into account the accuracy of the CNN without any consideration for embedded deployment or inference time.\par
Recent works like \cite{late_1,late_2,late_3,late_4} use a multi-objective or joint reward function to reduce power consumption and/or inference time besides improving accuracy. NetAdapt \cite{rl_5} proposes an iterative process to compress a pre-trained CNN by reducing the number of channels and employing empirical measurements on a target platform. He et al. \cite{rl_6} employ AutoML for model compression by having an actor-critic agent learn the compression policy of a network with latency and accuracy as reward. Each agent's action represents the desired compression rate and structure. \par

Overall, all the works employing ML for DSE are bound to a specific platform and do not offer support for a wide range of heterogeneous platforms. Besides, they address the problem of improving DNN architectures or compression, but do not give any attention to primitive selection optimization. Further, our method is complementary to those implementing graph optimizations as a final-processing step for them. To the best of our knowledge, we are the first ones to apply an RL-based search for primitive selection optimization on multiple target platforms.

\section{BACKGROUND: INFERENCE OF DNN ON HETEROGENEOUS EMBEDDED DEVICES}
Deep Neural Networks (DNN) are composed of a set of layers in cascade, e.g. convolution, pooling, activation and fully connected, that transform an input into a set of features maps which can be classified, detected or recognized based on a score or probability function. Training of DNN involves both a forward pass to compute the final score function and a backward pass to learn the weights according to a loss function. In this work, we address the problem of improving inference efficiency, that is, the forward pass latency of a DNN after training, and its deployment for industrial applications on heterogeneous embedded devices.

\subsection{INFERENCE ENGINE OPTIMIZER}
We form part of an European collaboration to bring Deep Learning to any party who would like to take up DL solutions in an industrial environment \cite{bonseyes}. One of the main goals of the project is to reduce development time and to optimize deployment of DNN on embedded systems. In this context, a neural network framework has been developed to produce efficient and tunable code which enables and maximizes the portability among heterogeneous platforms \cite{quenn}. The core of the inference engine optimizer comprises a set of CPU dependency-free functions which can be complemented by specific-platform acceleration libraries to generate optimized implementation for the system. In this work, we address the integration of the inference engine optimizer into our search environment to tightly couple empirical measurements of a heterogeneous platform to a learning-based search. \par

\subsection{ACCELERATION LIBRARIES} We present the set of libraries and primitives available in the inference engine optimizer for DNNs:
\begin{itemize}
\item \textbf{Vanilla:} This group embraces the set of CPU dependency-free and direct functions implemented in ANSI C with the objective of maximizing portability. It does not rely on any acceleration library.
\item \textbf{Basic Linear Algebra Subprograms (BLAS):} This group includes ATLAS and openBLAS libraries which implement GEMM and GEMV routines \cite{gemm} on CPU cores. Any of these libraries can use the following lowering methods: im2col, im2row and kn2row.
\item \textbf{NNPACK:} It is an open-source acceleration library which provides low-level performance primitives on CPU cores for specific DL layers \cite{nnpack}.
\item \textbf{ArmCL:} Set of high-performance routines for Arm processors. We have used Winograd transformation and BLAS routines for convolutional layers and specific-optimized code for Depth-Wise convolutions \cite{armcl}. 
\item \textbf{Sparse:} It includes multiple implementations which can be used to compress the model representation in memory for convolutional and FC layers.
\item \textbf{cuDNN:} Highly optimized primitives for Nvidia GPUs which implement several DNN routines \cite{cudnn}. It is important to remark that this library does not include a specific implementation for FC layer.
\item \textbf{cuBLAS:} BLAS routines for Nvidia GPUs \cite{cublas}. We have only used the GEMV routine for FC layer.
\end{itemize}

\section{LEARNING-BASED SEARCH ENGINE}
In this section, we address the problem of primitive selection and we propose Reinforcement Learning as a solution.
\subsection{PROBLEM FORMULATION}
\begin{figure}[t!]
   \centering
   \includegraphics[width=0.45\textwidth, height=1in]{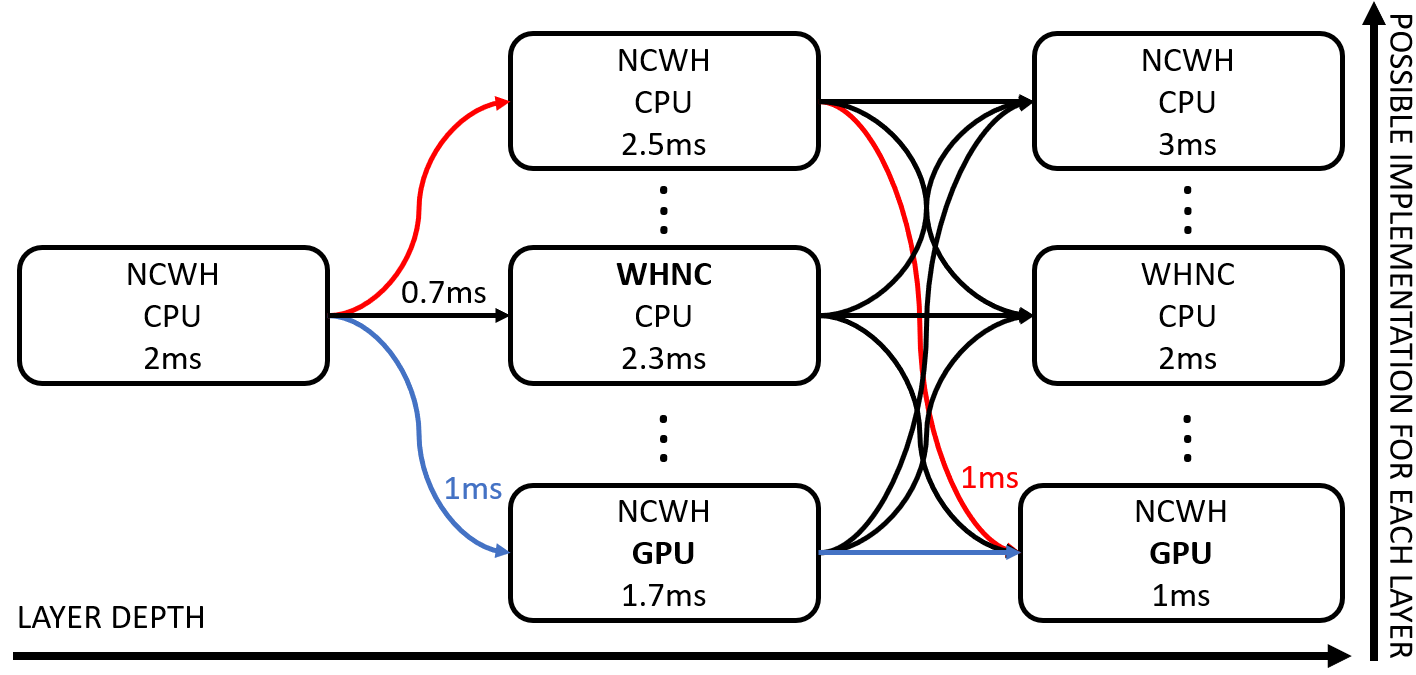}
   \caption{3-layer network. Arrows with time express incompatibility penalty. The agent is able to avoid local minimum, e.g. red path, which contains the fastest intermediate implementation. Instead, it selects the blue path: fastest overall.}
   \label{layers}
   \vspace{-0.1cm}
\end{figure}

Given a DNN, each layer may be executed by different acceleration libraries which, in turn, might provide several primitives to yield an optimal implementation. The problem is not as trivial as to benchmark all primitives individually and select the fastest for each layer to make up the optimal network implementation. Each primitive may have a different input or output tensor layout which might not correspond to those layouts of previous and following layers, e.g. NCWH and WHNC. Therefore, incompatibilities arise and a layout conversion layer is needed which incurs in a penalty. Likewise, in an heterogeneous environment, layers can be executed in different processor types which involves a costly (slow) memory transfer, see Fig. \ref{layers}. \par
The number of combinations within a network, which is the design space to explore, grows exponentially with the number of layers, $N_L$, having as base the number of different implementations for such layer, $N_I$. Hence, the design space size for a network would be 
\begin{math} 
	N_I^{N_L}
\end{math} as the worst case. 
It becomes a non trivial problem and therefore, a careful search must be carried out to select the right set of primitives that, combined among them and assuming the conversion penalties, yields the fastest inference. 

\subsection{REINFORCEMENT LEARNING}
Reinforcement Learning (RL) lends itself perfectly to exploring large design spaces due to its sample-based approach and far-sighted accumulative reward \cite{rl_1,rl_2}. Consider the network space exploration as a Markov Decision Process (MDP) containing an agent.
We are interested in learning a function that optimizes the agent's behavior i.e. mapping from state $s_t$ to actions $a_t$ without modeling the environment and only relying on the reward function. Q-learning \cite{ql} fits well this description as it is a model-free and value-based implementation, having the policy implicit in the value function. The action-value function is the expected return in a state $s_t$ taking an action $a_t$:
\begin{equation}
	q_\pi\left(s,a\right) = E_\pi\left[G_t | s_t = s, a_t = a\right]
\end{equation}
The objective of Q-learning is to maximize the total reward:
\begin{math}
R_T = \sum_{t=0}^{\infty}\gamma^{t}r_{t}
\end{math}
where $r_t$ is an individual reward and $\gamma$ is the discounted factor for successive states. Besides, Q-learning is an off-policy implementation, that is, it may follow a behavior policy $\mathcal{A}$ while targeting a greedy policy $\mathcal{B}$. Following Bellman's equation, we can iteratively update the action-value function as follows:
\begin{equation}
Q(s_t,a_t)=Q_{s_t, a_t}(1-\alpha) + \alpha\left[r_t + \gamma\max_{a}Q(s_{t+1},a)\right]
\label{bellman}
\end{equation}

\subsection{SEARCH ENGINE}
We consider an agent whose aim is to learn the optimal path among a large but finite set of states $\mathcal{S}$ i.e. layer representations, employing a set of actions $\mathcal{A}$ i.e. layer implementations. RL suits well the specifications of the problem that we address in this work. Inference time represents a clear reward function given by the environment that we aim to explore: a Deep Neural Network.  \par

The agent samples sequentially a new set of primitives for the network, layer by layer. The state space is defined as a tuple of the parameters that specify the execution of a layer with a certain primitive on a target platform, see table \ref{state_space}. All primitives are defined by an algorithm, its implementation format and a BLAS library. The agent chooses one primitive from the set of acceleration libraries given the current layer type. Based on the action, the agent moves to another state and the process is repeated until the end of the network.

\begin{table}[h]
\begin{center}
\begin{tabular}{|l||l|}
\hline
\hspace{0.2cm}State Parameters & \hspace{1.3cm} Definition \\
\hline
Layer type & Any layer e.g. convolution, pooling  \\
Layer depth & Position of the layer in the network\\
Acceleration Library & Name of the library \\
Algorithm & Routine type \\
Algorithm impl & Sub-routine or lowering method \\
Hardware processor & CPU, GPU, FPGA. \\
BLAS library & Library name \\
\hline
\end{tabular}
\end{center}
\caption{State Space. Parameters define the execution of a layer with a specific primitive on a target platform.}
\label{state_space}
\end{table}

Similar to Baker et al. \cite{rl}, we have implemented an $\epsilon$-greedy strategy \cite{rl_3} which trades off between exploitation and exploration. The agent starts mainly exploring the design space (random actions) to sample the diverse possibilities ($\epsilon=1$). We slowly decrease $\epsilon$ over the episodes for the agent to select the best actions and finally learn an optimal path: full exploitation ($\epsilon=0$). In addition, we have added an experience replay after each episode which helps the action-value function converge faster \cite{rl_4}. We have set the experience replay's buffer size to 128 following \cite{rl}.\par
Although initially we used the network inference time as unique reward signal, we have applied Reward Shaping for better convergence. The objective is to maximize the total reward, in this case, minimize the inference time. Hence, each state receives as reward its own layer inference time but reversing the sign, e.g. 0.01ms $\Rightarrow$ -0.01ms. Thanks to the Q-learning update rule, each layer also receives Q-knowledge from the best following state. Therefore, the agent is able to combine both sources of knowledge, look ahead and avoid local minima due to penalties introduced by incompatibility between layers, see Fig. \ref{layers}.

\section{Q-BASED SEARCH FOR DEEP NEURAL NETWORKS (QS-DNN)}
\begin{figure}[t!]
   \centering
   \includegraphics[width=0.45\textwidth, scale=0.4]{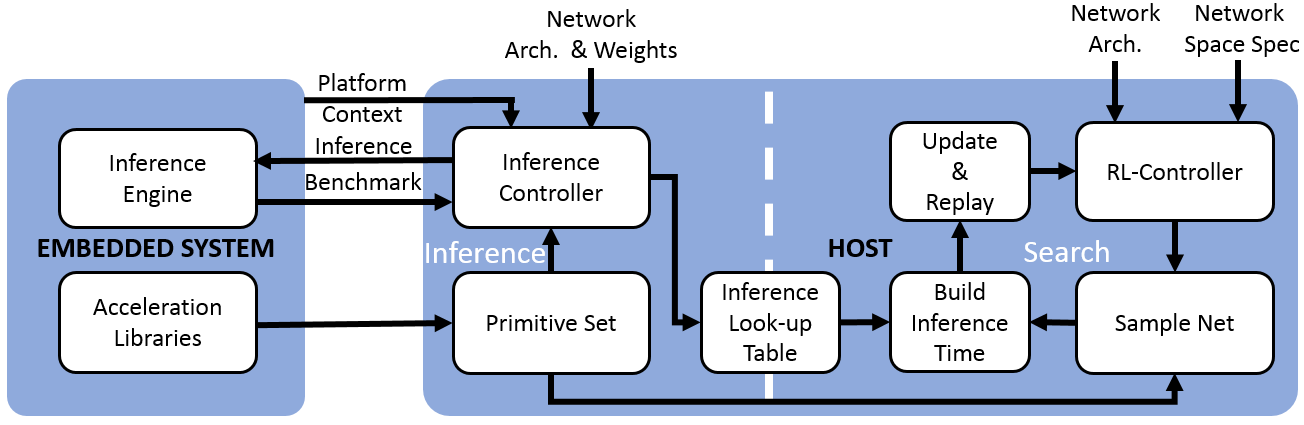}
   \caption{Architecture of QS-DNN. Complete flow: Inference on an embedded on the left, RL-based learning on the right.}
   \label{arch}
\end{figure}

The aim of QS-DNN is to automatically optimize the inference of any DNN and boost its performance on an embedded system. The process is composed of two phases: \textit{1)} inference of the DNN on the embedded system to obtain empirical measurements, \textit{2)} automatic RL-based search over a reduced number of episodes to explore the design space. We have separated the phases to avoid inferring on the embedded system each possible solution of the  space search, which would slow down remarkably the process, see Fig. \ref{arch}. \par

\begin{figure}[t!]
   \centering
   \includegraphics[width=0.45\textwidth, scale=0.3]{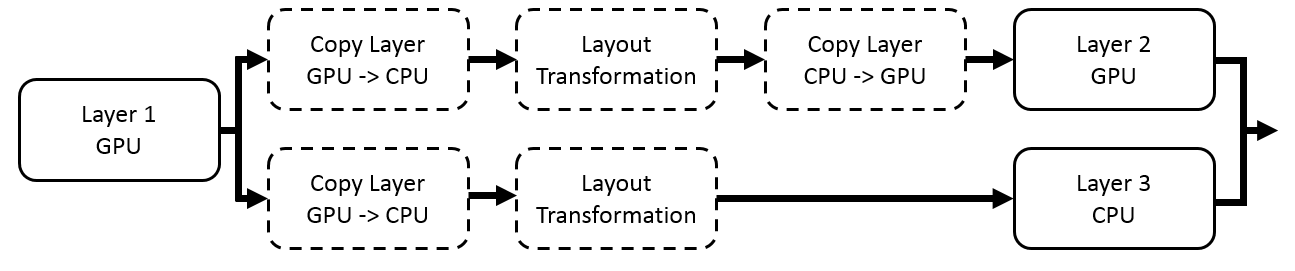}
   \caption{Profiling of compatibility layers between all consecutive layers. Exception and branches are handled.}
   \label{compatibility}
\end{figure}

\subsection{INFERENCE}
We employ the inference engine optimizer described in Section III and its acceleration libraries to obtain real measurements although the search could be also applied to any other inference framework. We consider Vanilla library as the base implementations for all measurements since it is the most simple, direct, dependency-free and it contains all layers that a DNN may use. \par

Having set the base, the inference controller benchmarks each primitive type\footnote{Each primitive is inferred for 50 images and the mean is calculated}, one at a time, by substituting Vanilla for the chosen primitive type in all those layers where the acceleration library is able to implement such primitive. Therefore, we only need to infer the whole network on the embedded platform as many times as different global implementations there exists. In each inference, the execution time for each layer is measured and retrieved. \par

Once all primitive types have been benchmarked, we profile the compatibility layers for layout transformation and data transfers between different processor. A single inference is performed to benchmark all possible compatibility layers between each consecutive layer of the neural network, see Fig.\ref{compatibility}. After all inference measurements have been retrieved, a look-up table is built.

\subsection{SEARCH}
The search space and the conditions of the search can be defined for each network. They specify the behavior of the agent: number of episodes for each $\epsilon$, learning rate, discounted factor and replay buffer's size. We have set the learning rate to 0.05 and discounted factor to 0.9 to give slightly more importance to short-term rewards. Once the inference phase has finished, the Q-learning -based search begins and proceeds as shown in Algorithm \ref{alg:QS-DNN}. 

\begin{algorithm}[t!]
\caption{QS-DNN - Search}\label{alg:QS-DNN}
\begin{algorithmic}[1]
\State $\epsilon \gets \epsilon_{new}$
\While{Learned Episodes $<$ Episodes($\epsilon$)}
\State Reset Path
\While{Layer $\neq$ End Layer($\epsilon$)}
\If{Generate Random $< \epsilon$}
  \State Action $\gets$ Q-values(Random)
\Else{}
  \State Action $\gets$ Q-values(Max)
\EndIf
\State Layer $\gets$ Next Layer
\EndWhile\label{euclidendwhile}
\State Check for Incompatibility
\State Compute Inference Time
\State Experience Replay \& Update (eq. \ref{bellman}) 
\EndWhile\label{layer}
\end{algorithmic}
\end{algorithm}

\begin{figure*}[t]
   \centering
   \includegraphics[width=0.9\textwidth, height=1.8in]{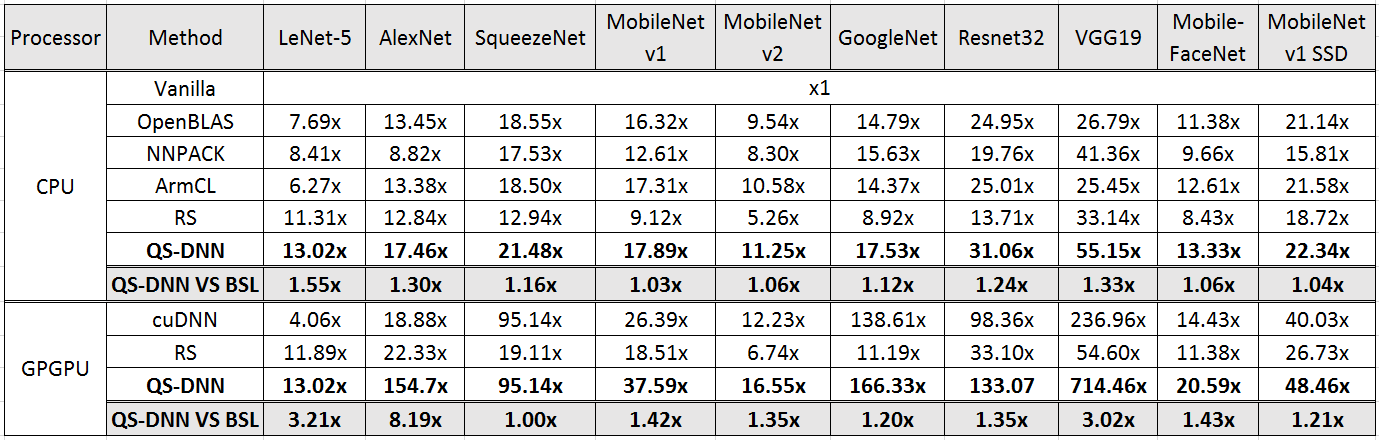}
   \captionof{table}{Inference time speedup of CPU- and GPGPU-based implementations respect to Vanilla (dependency-free implementation). Results correspond to most performing libraries employing their fastest primitive for single-thread and 32-bit floating-point operations. QS-DNN VS BSL shows the improvement of the search over the Best Single Library (BSL) and clearly outperforms RS (Random Search) for 1000 episodes.}
   \label{results}
\end{figure*}

\begin{figure}[t]
   \includegraphics[width=1\linewidth, height=1.2in]{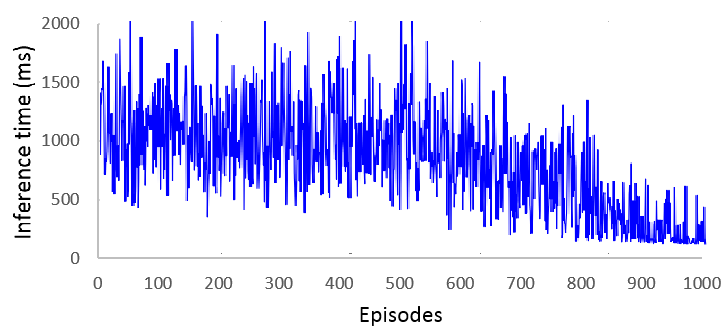}
  \caption{RL search for 1000 episodes where the 500 first episodes are fully exploration. From there on, $\epsilon$ is decreased by 0.1 towards exploitation after every 50 episodes.}
  \label{fig:sub1}
\end{figure}

First, $\epsilon$ is retrieved from the specifications as well as the number of episodes for such $\epsilon$. In all experiments, 50\% of the total episodes correspond to full exploration and 5\% to any other $\epsilon$ from 0.9 to 0.1. By these means, the agent obtains enough knowledge from the environment before starting exploitation, see Fig. \ref{fig:sub1}. \par

For each episode, the agent samples sequentially a new set of primitives based on the $\epsilon-$strategy. Once the network's configuration is set, the engine automatically looks for incompatibilities between layers due to layout and processor type. At last, the total network inference time is computed by looking up each implementation in the inference table and summing up the execution time of all layers. If any incompatibility has been found between two layers, the extra penalty is added to the inference time of the latter layer. Finally, the action-value function is updated with the current reward and stored for experience replay. When the number of episodes for a given $\epsilon$ has been met, $\epsilon$ is decreased towards exploitation phase. By the end of the search, the engine gives out the best inference configuration and the learning curve that the agent has followed, see Fig \ref{fig:sub1}.

\section{RESULTS AND DISCUSSION}
In this section, we show the results from applying QS-DNN to several DNNs for image classification, face recognition and object detection tasks. 
\subsection{INFERENCE OPTIMIZATION}
All inference experiments have been conducted on the heterogeneous platform Nvidia Jetson TX-2 using single-precision floating-point operations. All CPU inferences correspond to using a single-thread on an ARM Cortex A-57 core, while GPGPU inferences correspond to using either the single-thread CPU or the Nvidia Pascal GPU which features 256 cores. The design space search is carried out in a standard Intel CPU and takes less than 10 min. to converge. \par

Given the acceleration libraries from Section III, the maximum number of different primitive for a layer, taking all the variants, is 13. Table \ref{results} summarizes the results of the most performing implementations. It is possible to observe that, QS-DNN outperforms all single-library implementations and achieves considerable speedups compared to the Best Single Library (BSL) for CPU and GPGPU modes. \par 

It is interesting to note that the fastest implementation for Lenet-5 in GPGPU mode is actually a pure CPU implementation. In this case, the agent learns that, despite GPU implementation is faster for some layers, data transfers between CPU and GPU diminish the speedup that GPU yields. It is also possible to note a great improvement of QS-DNN (GPGPU) over cuDNN in VGG19 or AlexNet as cuDNN does not implement the costly FC layer of these networks. In particular, QS-DNN (GPGPU) achieves a notable speedup for MobileNets (over 1.4x) where it learns to combine the optimized Depth-Wise code from ArmCL, convolutions from cuDNN and certain ReLU and B-Norm layers from Vanilla to avoid costly extra copies to GPU.

\subsection{REINFORCEMENT LEARNING VS RANDOM SEARCH}
In this section, we address the learning process of RL and compare it to Random Search (RS). RL outperforms RS in all networks and achieves speedups of up to x15 over RS for larger design spaces, e.g. GoogleNet or VGG19, on GPGPU mode, see Table \ref{results}. 
\begin{figure}[ht]
\centering
   \includegraphics[width=1\linewidth, height=1.2in]{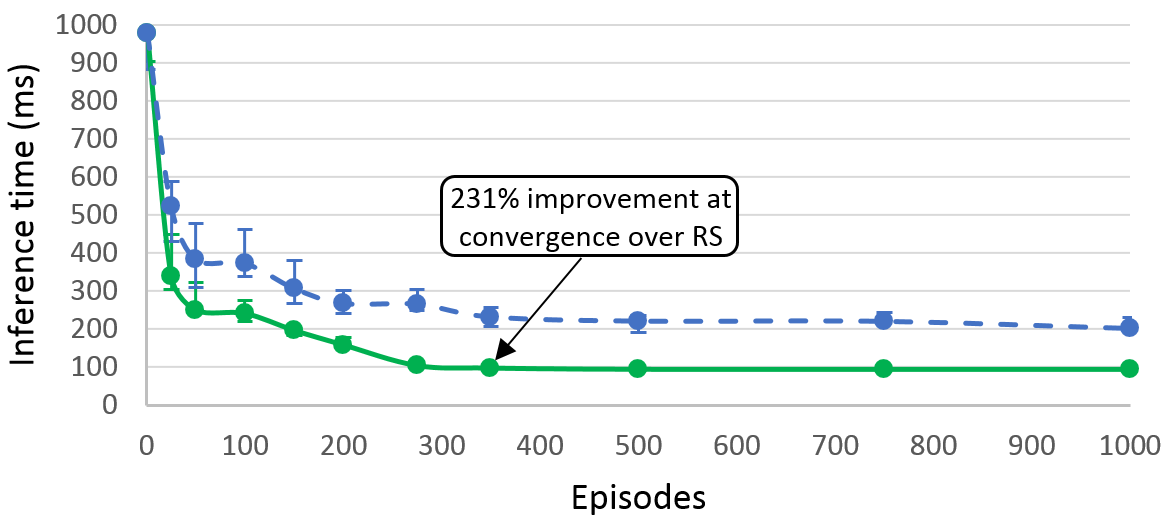}
  \caption{RL VS RS for Mobilenet. Each point indicates the average result for a complete search for the given episodes. Variance reduces towards the end as the search converges.}
  \label{fig:sub2}
\end{figure}

Fig. \ref{fig:sub2} gives an example of RL VS RS for MobileNet-v1 where each point represents the mean inference time from 5 full experiments for a reduced budget: number of episodes. With a budget of a few episodes, the variance of both implementation is high as they do not obtain much knowledge about the environment. RL's solutions quickly decrease inference time as the agent observes more episodes and it falls near convergence after only 350. On the other hand, RS fails to find implementations as optimized as RL's since it does not implement any learning method. RS's solutions are already 50\% worse than RL's with only 25 episodes and twice as worse after 350 episodes. RS's implementations decrease inference time after seeing more options as it discards naive implementations, but it only converges towards the infinite. 

\section{CONCLUSIONS AND FUTURE WORK}
We have presented an automatic exploration framework which relies on a design space search based on Reinforcement Learning (RL). The RL-based search efficiently learns an optimized combination of primitives to tune and boost the inference of DNNs. The search is tightly coupled with an inference engine optimizer which facilitates the deployment and optimization of DNNs on heterogeneous embedded platforms.
We have shown that, the search, together with the inference engine optimizer, is able to achieve 2x speedup on average in inference latency compared to the best single vendor library in a GPGPU platform. Further, the RL-based search quickly converges and outperforms Random Search achieving up to 15x better results in large design spaces. In addition, our approach is very modular and can be applied to other optimization methods as a post-processing step.\par
We aim to extend this work to other heterogeneous target platforms, e.g FPGA, VPU or ASIC\footnote{If API at network-layer level is provided e.g. Convolution}. In addition, we envision to extend exploration to e.g. different reward choices or having multi-objective search, for problems related to inference of DNNs on constrained environments. Further, we also aim to look into Deep RL to approximate the value function for better scalability towards larger networks and more dimensions in the search space.

\addtolength{\textheight}{-12cm}   




\section*{ACKNOWLEDGMENT}
{\small This project has received funding from the European Union’s Horizon 2020 research and innovation programme under grant agreement No. 732204 (Bonseyes). This work is supported by the Swiss State Secretariat for Education‚ Research and Innovation (SERI) under contract number 16.0159. The opinions expressed and arguments employed herein do not necessarily reflect the official views of these funding bodies.}


\begin{thebibliography}{99}

\bibitem{google} Google AI. URL: https://ai.google/
\bibitem{tesla} Tesla. URL: https://www.forbes.com/sites/bernardmarr/2018/01/08/the-amazing-ways-tesla-is-using-artificial-intelligence-and-big-data
\bibitem{facerecognition} Rowley, Henry A., Shumeet Baluja, and Takeo Kanade. "Neural network-based face detection." IEEE Transactions on pattern analysis and machine intelligence 20.1 (1998): 23-38.
\bibitem{speechrecognition} Graves, Alex, Abdel-rahman Mohamed, and Geoffrey Hinton. "Speech recognition with deep recurrent neural networks." Acoustics, speech and signal processing (icassp), 2013 ieee international conference on. IEEE, 2013.
\bibitem{imagenet} Imagenet. URL: http://www.image-net.org.
\bibitem{anoverview} Shafique, Muhammad, et al. "An overview of next-generation architectures for machine learning: Roadmap, opportunities and challenges in the IoT era." Design, Automation \& Test in Europe Conference \& Exhibition (DATE), 2018. IEEE, 2018.
\bibitem{caffe} Jia, Yangqing, et al. "Caffe: Convolutional architecture for fast feature embedding." Proceedings of the 22nd ACM international conference on Multimedia. ACM, 2014.
\bibitem{tensorflow} Abadi, Martín, et al. "Tensorflow: a system for large-scale machine learning." OSDI. Vol. 16. 2016.
\bibitem{cudnn} Chetlur, Sharan, et al. "cudnn: Efficient primitives for deep learning." arXiv preprint arXiv:1410.0759 (2014).
\bibitem{armcl} Arm Compute Library. URL: https://developer.arm.com-technologies/compute-library
\bibitem{qualcomm} Qualcomm. URL: https://www.qualcomm.com/snapdragon
\bibitem{xilinx} Xilinx. URL: https://www.xilinx.com/
\bibitem{nvidia} https://www.nvidia.com/en-us/
\bibitem{andrew} Anderson, Andrew, and David Gregg. "Optimal DNN primitive selection with partitioned boolean quadratic programming." arXiv preprint arXiv:1710.01079 (2017).
\bibitem{processor} Ozisikyilmaz, Berkin, Gokhan Memik, and Alok Choudhary. "Efficient system design space exploration using machine learning techniques." Proceedings of the 45th annual design automation conference. ACM.
\bibitem{synthesis} Liu, Hung-Yi, and Luca P. Carloni. "On learning-based methods for design-space exploration with high-level synthesis." Proceedings of the 50th annual design automation conference. ACM, 2013. 
\bibitem{ea_1} Real, Esteban, et al. "Regularized evolution for image classifier architecture search." arXiv preprint arXiv:1802.01548 (2018).
\bibitem{ea_2} Cortes, Corinna, et al. "Adanet: Adaptive structural learning of artificial neural networks." arXiv preprint arXiv:1607.01097 (2016).
\bibitem{ea_3} Al-Hyari, Abeer, and Shawki Areibi. "Design space exploration of Convolutional Neural Networks based on Evolutionary Algorithms." Journal of Computational Vision and Imaging Systems 3.1 (2017).
\bibitem{late_1} Hsu, Chi-Hung, et al. "MONAS: Multi-Objective Neural Architecture Search using Reinforcement Learning." arXiv preprint arXiv:1806.10332 (2018).
\bibitem{late_2} Tan, Mingxing, et al. "MnasNet: Platform-Aware Neural Architecture Search for Mobile." arXiv preprint arXiv:1807.11626 (2018).
\bibitem{late_3} Dong, Jin-Dong, et al. "PPP-Net: Platform-aware Progressive Search for Pareto-optimal Neural Architectures." (2018).
\bibitem{late_4} Kim, Ye-Hoon, et al. "Nemo: Neuro-evolution with multiobjective optimization of deep neural network for speed and accuracy." ICML.
\bibitem{boda} Moskewicz, Matthew W., Ali Jannesari, and Kurt Keutzer. "Boda: A Holistic Approach for Implementing Neural Network Computations." Proceedings of the Computing Frontiers Conference. ACM, 2017.
\bibitem{latte} Truong, Leonard, et al. "Latte: a language, compiler, and runtime for elegant and efficient deep neural networks." ACM SIGPLAN Notices 51.6 (2016): 209-223.
\bibitem{nnpack} NNPACK. URL: https://github.com/Maratyszcza/NNPACK
\bibitem{cublas} cuBLAS. URL: https://docs.nvidia.com/cuda/cublas/index.html
\bibitem{gemm} BLAS: URL: http://www.netlib.org/blas/
\bibitem{rl} Baker, Bowen, et al. "Designing neural network architectures using reinforcement learning." arXiv preprint arXiv:1611.02167 (2016).
\bibitem{rl_1} Li, Yuxi. "Deep reinforcement learning: An overview." arXiv preprint arXiv:1701.07274 (2017).
\bibitem{rl_2} Sutton, Richard S., and Andrew G. Barto. Introduction to reinforcement learning. Vol. 135. Cambridge: MIT press, 1998.
\bibitem{ql} Watkins, Christopher John Cornish Hellaby. Learning from delayed rewards. Diss. King's College, Cambridge, 1989.
\bibitem{rl_3} Mnih, Volodymyr, et al. "Human-level control through deep reinforcement learning." Nature 518.7540 (2015): 529.
\bibitem{rl_4} Lin, Long-Ji. "Self-improving reactive agents based on reinforcement learning, planning and teaching." Machine learning 8.3-4 (1992).
\bibitem{rl_5} Yang, Tien-Ju, et al. "NetAdapt: Platform-Aware Neural Network Adaptation for Mobile Applications." arXiv preprint arXiv:1804.03230
\bibitem{rl_6} He, Yihui, and Song Han. "ADC: Automated Deep Compression and Acceleration with Reinforcement Learning." arXiv preprint arXiv:1802.03494 (2018).
\bibitem{quenn} de Prado, Miguel, et al. "QUENN: QUantization engine for low-power neural networks." Proceedings of the 15th ACM International Conference on Computing Frontiers. ACM, 2018.
\bibitem{bonseyes} Llewellynn, Tim, et al. "BONSEYES: platform for open development of systems of artificial intelligence." Proceedings of the Computing Frontiers Conference. ACM, 2017.

\end{thebibliography}
\end{document}